# The Grievance Dictionary: Understanding Threatening Language Use

Isabelle van der Vegt[1], Maximilian Mozes[1], Bennett Kleinberg[1,2] and Paul Gill[1]
[1] Department of Security and Crime Science, University College London
[2] Dawes Centre for Future Crime, University College London
Correspondence: isabelle.vandervegt@ucl.ac.uk

**Abstract**
This paper introduces the Grievance Dictionary, a psycholinguistic dictionary which can be used to automatically understand language use in the context of grievance-fuelled violence threat assessment. We describe the development the dictionary, which was informed by suggestions from experienced threat assessment practitioners. These suggestions and subsequent human and computational word list generation resulted in a dictionary of 20,502 words annotated by 2,318 participants. The dictionary was validated by applying it to texts written by violent and non-violent individuals, showing strong evidence for a difference between populations in several dictionary categories. Further classification tasks showed promising performance, but future improvements are still needed. Finally, we provide instructions and suggestions for the use of the Grievance Dictionary by security professionals and (violence) researchers.

**Introduction**
Psycholinguistic dictionaries assume language use reflects the emotions and cognitive processes of a text author (Pennebaker et al., 2015; Pennebaker & King, 1999). Consequently, these processes are thought to be measurable, for example by examining a text for words that refer to a specific process or concept. One of the most prominent examples of a word-count based psycholinguistic dictionary is the Linguistic Inquiry and Word Count (LIWC; Pennebaker et al., 2015). It provides a "method for studying the various emotional, cognitive, and structural components present in individuals' verbal and written speech samples" (Pennebaker et al., 2015, p.1). In short, LIWC seeks to measure variables relating to linguistic style (e.g. word count, pronouns, number of verbs), psychological processes (e.g. anxiety, power), and personal concerns (e.g. family, religion). Others dictionaries (e.g. Wmatrix, Rayson, 2008; Empath, Fast et al., 2016; Moral Foundations Dictionary, Frimer et al., 2019; IBM Watson Tone Analyzer[1]) also exist and measure different concepts and categories, however LIWC remains pre-eminent within research circles.

Such psycholinguistic dictionaries are increasingly used to understand and detect extreme, threatening or hateful language on the web (Davidson et al., 2017; Kleinberg, van der Vegt, & Gill, 2020; Scrivens et al., 2018). They are also used to inform automatic linguistic threat assessment (e.g. Akrami et al., 2018). Threat assessments can cover a range of threats of violence including violent extremism, public mass murder, school shootings, and targeted violence against public figures. These forms of violence share a similar genesis, typically involve some form of pre-planning, and are driven by a grievance (Corner et al., 2018). They are also often signalled ahead of time in some form of written communication (Gill, 2020). Typically, research on automatic linguistic threat assessment tries to discriminate between texts authored by perpetrators of grievance-fuelled violence from some form of non-violent control or comparison group (Baele, 2017; Kaati, Shrestha, & Cohen, 2016).

---

[1] https://www.ibm.com/watson/services/tone-analyzer/

However, the psycholinguistic dictionaries frequently used in these contexts are met with two important limitations. Firstly, standard psycholinguistic dictionaries have not been developed for the purpose of assessing grievance-fuelled language and therefore do not measure constructs that may be of interest to researchers and threat assessment practitioners. Although the LIWC provides categories such as anxiety and anger, we propose that key concepts for threat assessment and violence research are absent in this and other dictionaries. As a result, previous work on grievance-fuelled violence that used the LIWC (e.g., Baele, 2017; Kaati, Shrestha, & Cohen, 2016) may not have used valid measures of potential violence. Second, the content and construction procedure of existing dictionaries is often unclear, because descriptions of how and why certain words have been selected are scarce. Yet, it is vital to be transparent about the development of these because of the far-reaching consequences of false positives and negatives within the context of threat assessment. In the UK, the ALGO-CARE framework suggests that algorithms used in the context of policing need to be explainable, in that decision-making rules and the impact of each factor on the outcome is available (Oswald et al., 2018). In short, it is highly important for practitioners and researchers to understand the capabilities and limitations of a given dictionary. For many available dictionaries and threat assessment software[2], the contents of wordlists or other 'under the hood' operations are not available to its users, and thus cannot be adequately evaluated or explained. This possibility is desirable and necessary if such systems are to be used in practice.

To address the aforementioned limitations, this paper presents the Grievance Dictionary, which specifically aims to measure psychological and social concepts that are of interest in the context of grievance-fuelled violence threat assessment. First, the Grievance Dictionary is specifically aimed at measuring concepts that are of interest in threat assessment and violence research and practice. Its aim is to supplement measures obtained through dictionaries such as the LIWC with concepts that are specifically relevant to the threat assessment domain. Second, the Grievance Dictionary is transparent in terms of its construction and final format. All data collected are made available freely (e.g., for researchers and practitioners), including the words that are included in the final dictionary as well as background characteristics of consulted experts. Third, the dictionary is not restricted to a specific type of violence or extremism. Any threat, abuse, or violent writing fuelled by a grievance can be assessed with the Grievance Dictionary. This would apply to a wide spectrum of phenomena, including right and left-wing extremism, religious extremism, and (in many cases) threats directed at public officials. Resultingly, dictionary terms will not necessarily need to be continuously updated as is the case for other domain-specific dictionaries.

In the following section, we discuss psycholinguistic dictionaries and their use in threat assessment. In part one, we discuss how the Grievance Dictionary was developed through expert consultation, human and computational word list generation, and crowdsourced annotations. We also perform a psychometric evaluation for each dictionary category. In part two, we present empirical results using the final dictionary. The dictionary is validated by performing statistical comparisons as well as classification taks on several datasets. We conclude with a general discussion of the dictionary development and validation, as well as possible future avenues.

---

[2] This includes endeavors such as PRAT (Akrami et al., 2018) and https://threattriage.com/

**Automatic linguistic threat assessment and grievance-fuelled violence**
In the automatic linguistic approach to grievance-fuelled violence, particular attention has been paid to the writings of terrorists and (online) extremists (Baele, 2017; Kaati, Shrestha, & Cohen, 2016; Kop et al., 2019). A few studies examined lone-actor terrorist manifestos for various psycholinguistic variables using the LIWC (Baele, 2017; Kaati et al., 2016). These studies compared lone-actor terrorist writings to the writings of several different populations, such as non-violent activists (e.g. Martin Luther King, Nelson Mandela), standard control writings and emotional writings (i.e., 'baseline' texts expressing low and high emotionality, respectively), and personal blogs (Kaati et al., 2016). In several studies, lone-actor terrorist manifestos differed from control texts on several LIWC variables. For example, they contained higher proportions of negative emotion words including anger (Baele, 2017; Kaati et al., 2016), lower levels of positive emotion and friendship words (Kaati et al., 2016), and more power-related words (Baele, 2017; Kaati et al., 2016). Similar research focused on 'incel' (i.e. involuntary celibate) forums. Jaki et al (2019) compared 50,000 messages from an incel forum to 50,000 neutral 'control' texts extracted from Wikipedia articles and random English tweets via LIWC software. Incel messages contained more swear words and negative emotion, such as anger and anxiety.

Besides the use of the LIWC, several studies in extremism additionally use custom-made 'expert dictionaries'. For these dictionaries, domain experts are consulted to develop wordlists that cover the terms used by a specific population. For example, Smith et al. (2020) developed an expert dictionary for ISIS vernacular after consulting 'terrorism and extremism experts from government and the security and defence sectors' (Smith et al., 2020, p.6). Figéa et al. (2016) developed one for racism, aggression, and worries on a white supremacy forum.

Using the LIWC as well as expert dictionaries, other studies go beyond statistical comparisons alone, and classify violent from non-violent texts via machine learning. In a study of a white supremacy forum, all 73 LIWC categories and three expert dictionaries relating to worries, racism and aggression were used as features (Figea et al., 2016). LIWC categories for religion (e.g. 'Muslim', 'church'), see (e.g. 'view', 'saw') and third person pronouns (e.g. 'they', 'them') proved important linguistic characteristics for classifying racism posts. The LIWC categories for anger (e.g. 'hate', 'kill') and an expert dictionary category for aggression were important for recognizing both worries and aggression in the posts, achieving accuracy rates between 80-93%. In another effort, classification tasks using the LIWC output as predictors distinguished between lone-actor terrorist manifestos, texts written by non-violent activists, texts from personal blogs, forum postings on Stormfront (a white supremacy forum), and personal interest forum postings (Kaati, Shrestha, & Sardella, 2016). In one of the tasks where the aim was to distinguish between terrorist texts and Stormfront posts, LIWC categories relating to negative emotion (e.g. 'sad', 'angry'), time (e.g. 'before', 'often'), and seeing (e.g. 'appear', 'show') were important features for classification with an accuracy of 90%.

In earlier work, the concepts of hate and violence were measured on American and Middle Eastern dark web forums (Abbasi & Chen, 2007). The authors utilised a custom dictionary containing words and phrases from the forums related to violence and hate (the content of the dictionary was not made available). The results indicated that Middle Eastern forums scored higher than American forums in terms of violence. Forums from both regions did not differ in terms of hate. Similarly, Chen (2008) proposed an automated method for analysing affect within two jihadist dark web forums. Up to 909,039 messages were collected from the forums, of which 500 were utilised to manually construct a dictionary for violence, anger, hate and racism. One of the forums, known to be more radical, was indeed found to contain higher levels of violence, anger, hate, and racism than the other (Chen, 2008).

Custom dictionaries created through expert consultation also potentially suffer from a third limitation in addition to the two noted in the introduction. They are often highly domain-dependent and non-transparent regarding the population of experts consulted. By consulting domain experts (e.g., in right-wing extremism, radical Islam) the dictionaries are specifically attuned to a specific type of violence or extremism. The nature of online communication in these populations is that language is community-specific and constantly changes (Farrell et al., 2020; Shrestha et al., 2017). Some fringe communities may also continuously adapt their language use to evade content moderation filters on social media platforms which automatically delete or flag posts with specific word use (van der Vegt, Gill, et al., 2019). As a result of these phenomena, dictionaries would have to be continuously updated to capture the appropriate jargon. Furthermore, custom expert dictionaries are referenced in Abbasi & Chen (2007), Chen (2008), Figea et al. (2016) and Smith et al. (2020), but little is said about what the consultation process entailed and why those consulted can be considered experts. In short, readers are expected to trust the judgment of the researchers and experts without having access to the specifications of the tool.

*Transparency statement*
The approach to developing the Grievance Dictionary was fully pre-registered before data collection: https://osf.io/szvm7. All data and materials are available on the Open Science Framework: https://osf.io/3grd6/. A user guide for the dictionary can be found there too.

**The Grievance Dictionary**

**Part I: Dictionary development**

The dictionary development consisted of five phases. (1) Threat assessment experts suggested dictionary categories. (2) Human subjects generated seed terms for each category. (3) Computational linguistics methods augmented the word list. (4) Human annotators rated word candidates on their fit into a set of categories. (5) The internal reliability for each dictionary category is assessed and their correlation with LIWC2015 categories is computed.

*Phase 1: Expert survey*
An online survey was sent out to experts within the field of threat assessment. Participants were professional contacts of the involved researchers in the field of threat assessment and terrorism research. Participants were asked the following:

*Imagine you are tasked with assessing whether a piece of text signals a threat to commit violence against a designated area, individual, or entity. It may be a physical letter or an online message that you are asked to examine. In short, you are trying to judge whether the person who wrote the text will act on their threat.* **What do you look for in the text to assess its threat level?** *Please mention all relevant factors that come to mind.*

The response to this question was an open text box, with no word limit. Following this, participants could add any other relevant factors that came to mind (again with an open answer response) and were asked about their professional experience in threat assessment (in years) and with linguistic threat assessment (on a 10-point scale, 1 = no experience, 10 = a lot of experience).

In total, 21 responses were gathered. On average the participants had 16 years of experience with threat assessment (SD = 8.84, range: 2-30 years). Overall, the participants indicated they had significant experience with threat assessment based on language, with a mean score of 8.17 (SD = 2.04, on a scale from 1-10).

Based on the survey responses gathered, it became clear that assessing the threat of violence through language relies on a wide variety of factors. In order to adequately measure these factors, they need to be condensed into psycholinguistic categories (e.g., similar to the LIWC). The lead author categorised free text responses. For example, the concepts 'preparation', 'rehearsal', 'developing capacity', 'refining method', or 'developing opportunity', were all coded as a single category relating to 'planning'. In total, this resulted in 79 categories (available on the OSF). The categories could broadly be defined to relate to the content of a communication (e.g., direct threat, violence, relationship), emotional processes (e.g., anger, frustration, desperation), mental health aspects (e.g., psychosis, delusional jealousy, paranoia), the communication style (e.g., unusual grammar, politeness, incoherence), and meta-linguistic factors (e.g., number of communications, font, use of graphics). Lastly, the lead author selected categories that could feasibly be represented as a psycholinguistic wordlist, serving as an overarching category (e.g., including 'weaponry' but excluding 'mentioning target' because it is too situation-specific). This resulted in a final selection of 22 categories (Table 1).

Table 1. Dictionary categories with example words (defined in later steps)

| Category | Examples | Category | Examples |
|----------|----------|----------|----------|
| Planning | long-term, tactic, organise | Deadline | time run out, due date, upcoming |
| Violence | bloodshed, fight, bullet | Murder | kill, stab, fatal |
| Weaponry | AK-47, ammo, fire arm | Relationship | marry, romantic, love |
| Help seeking | support, SOS, save | Loneliness | disconnected, nobody, abandon |
| Hate | enemy, loathe, hatred | Surveillance | spy, CCTV, monitor |
| Frustration | annoyed, problem, powerless | Soldier | fighter, battle, patriot |
| Suicide | die, overdose, last resort | Honour | integrity, hero, brave |
| Threat | warn, danger, unsafe | Impostor | impersonate, fraudulent, undercover |
| Grievance | wrong, disappointed, injustice | Jealousy | cheat, resent, bitter |
| Fixation | obsess, possess, watch | God | pray, holy, almighty |
| Desperation | sorrow, last chance, urgent | Paranoia | suspicious, conspiracy, suspect |

*Phase 2: Seed word generation*

Human subjects generated seed words for each category from Phase 1. A total of 13 participants suggested words for the categories in an online survey. Participants were all PhD students at English-speaking universities (full details of the sample are reported in the supplementary materials on OSF). For each category, participants were asked to write down all the words that came to mind, considering the category as an over-arching concept for the words they noted down. This resulted in a total of 1,951 seed words across categories. Instructions for the word generation task as well as the resulting words for each category are available in the online materials.

*Phase 3: Word list extension*

Two processes extended the word list. First, WordNet (Fellbaum, 1998) provided semantic associations for each seed word. This tool provides a lexical database of English words, grouped into 'cognitive synonyms' of meaningfully related words, which are added to the wordlists (e.g. 'knife' is supplemented with 'dagger', 'machete', and 'shiv'). All words related to the initial seed words were added to the list of the respective category.

Second, we obtained pre-trained word embeddings for each candidate word using GloVe, an unsupervised learning approach trained on a 6 billion word corpus (Pennington et al., 2014). GloVe represents words as real-valued vectors (embeddings) which aim to encode semantic relationships between individual words based on the contexts in which they appear.

This means that words which are similar in meaning have vector representations that are close to each other (based on a similarity measure) in the resulting vector space (e.g. a word embedding for 'gun' appears close to 'handgun', 'pistol', 'firearm', etc. in the learned vector space). For the dictionary, each seed word across all categories was supplemented with its ten nearest neighbour words in terms of cosine similarity. After removing duplicates obtained through WordNet and the embeddings, the final resulting wordlist across all categories contained 24,322 words. It is important to note that some words may appear in multiple categories (e.g. 'knife' may appear in both the weaponry and murder category).

*Development phase 4a: Word list rating*
Human annotators rated all words obtained through Phase 3 for the extent to which they fit within their respective category. An online task was developed where participants were presented with a category, a word, and the option to select, on a scale, 'how well the word displayed fits into the above category' (0 = does not fit at all, 10 = fits perfectly). They also had the option to select 'I do not know this word'. After reading instructions and consenting to participating, a total of 100 words (i.e., a random sample of 100 word-category pairs, with words shown for their associated category only) were rated by each participant. Participants were recruited through the crowdsourcing platform Prolific Academic, and remunerated for their time. Human workers were only eligible to participate if their first language was English. Interspersed between normal items, four attention checks were included (e.g. 'This is an attention check. Rate this word with 9 to continue').

In sum, the 24,322 words of the extended wordlist were rated by 2,318 online participants. A total of 238,366 ratings were obtained, with each word receiving at least 7 ratings, with an average of 9.42 ratings per word. All ratings from participants who failed at least one of the attention checks were removed (1.81%). Words for which the majority (50% or more) of participants indicated that they did not know the word, were also removed from the dictionary (0.39%). Following this, all dictionary words were stemmed and the ratings averaged per word stem (e.g. the ratings for 'friendship', 'friendly', and 'friends' were combined into a single score for the stem 'friend-'). This resulted in a final list of 20,502 words.

*Development phase 4b: Scoring methods*
Departing from the rated word list, several versions of the Grievance Dictionary can be used. Three possibilities are discussed. The first two rely on proportional scoring, based on word counts. Following the LIWC, we may wish to only retain words which received a high rating for belonging to a specific category (Pennebaker et al., 2015). In this first version, we would retain only those words which received an average rating of 7 or higher, resulting in a dictionary with 3,643 words. This version is used for evaluation and validation in this paper. An alternative second version retains words with a score of 5 or higher, resulting in a dictionary with 7,588 words. In both of these versions, scoring the texts follows the same approach as the LIWC, which is based on word count. When the dictionary is applied to a text, each word in the dictionary is searched and a proportion score for the word (i.e. frequency of the word / all words in text) and the overall category (i.e. frequency of all words in category / all words in text) is reported.

The third approach relies on average scoring, using the ratings assigned to each word through crowdsourcing. This version of the dictionary makes use of all 20,502 words and their associated average rating, assigning each word match in a text the appropriate weight. To measure each category for a text of interest, the average weight of all word matches per category is reported. While the first version using proportional scoring of words with a mean score of 7 and higher is used in this paper, alternative versions are available on the Open Science Framework.

*Development phase 5: Psychometric dictionary evaluation*

To assess the quality of the dictionary, it is important to examine the internal consistency of each category by measuring whether the words in each category yield a similar score for the respective category. We compute Cronbach's alpha using the proportional occurrence of each word in the 22 categories for a total of 17,583 texts across four corpora (Table 2). Similar to the development of LIWC2015 we use a varied selection of texts to compute reliability, including texts from deception detection experiments (Kleinberg et al., 2019), novels (Lahiri, 2014), movie reviews (Maas et al., 2011), and Reddit posts (Demszky et al., 2020).

When assessing the reliability of psychological tests, typically a Cronbach's alpha score of 0.70 or higher is considered acceptable (Taber, 2018). Cronbach's alpha ranges between 0 to 1 and is based on the covariance between items, where a score of 1 represents perfect covariance, such that the items adequately measure the same underlying concept. As raised in Pennebaker et al. (2015), assessing the reliability of dictionaries is somewhat more complicated. In language, similar concepts are typically not repeated several times; once something has been said it is generally not necessary to be said again. In contrast, similar concepts may be assessed repeatedly in psychological test items. Thus, it has been argued that an acceptable alpha score for dictionary categories will be lower than that for a psychological test (Pennebaker et al., 2015).

Table 2. Corpora used for internal consistency computation

| Corpus | Number of texts (number of tokens) |
|---|---|
| Deception detection experiments* | 2,547 (454,217) |
| Novels (Lahiri, 2014) | 3,036 (247,142,420) |
| IMDB movie reviews (Maas et al., 2011) | 50,000 (13,934,687) |
| Reddit posts (Demszky et al., 2020) | 70,000 (1,081,539) |

*Note.* *Hotel reviews (Ott et al., 2011, 2013), descriptions of past and planned activities (Kleinberg et al., 2019)

A psychometric evaluation was performed for each version of the dictionary (words with a rating of 7 or higher, words with a rating of 5 or higher, weighted words). The results reported from here onwards concern the dictionary using words with a rating of 7 or higher, because this dictionary performed best (results for the other versions are available on the OSF). The average alpha scores across corpora are reported in Table 3. The highest reliability of 0.41 is achieved for the category 'soldier', followed by 0.40 for 'god'. The lowest score (0.15) was found for the category 'grievance', which possibly shows that this concept is difficult to reliably measure with the current approach. The average reliability achieved across categories was 0.30 ($SD = 0.08$). This average reliability is close to the average reliability of 0.34 achieved with the LIWC 2015. The alpha scores for the LIWC2015 ranged between 0.04 and 0.69, whereas ours range between 0.20 to 0.39.

Table 3. Internal consistency scores

| Category | Cronbach's alpha | Category | Cronbach's alpha |
|---|---|---|---|
| deadline | 0.30 | loneliness | 0.21 |
| desperation | 0.21 | murder | 0.38 |
| fixation | 0.18 | paranoia | 0.29 |
| frustration | 0.28 | planning | 0.35 |
| god | 0.40 | relationship | 0.38 |
| grievance | 0.15 | soldier | 0.41 |
| hate | 0.34 | suicide | 0.27 |
| help | 0.24 | surveillance | 0.31 |
| honour | 0.30 | threat | 0.39 |
| impostor | 0.26 | violence | 0.39 |
| jealousy | 0.21 | weaponry | 0.38 |

In addition to internal reliability, we also assessed whether and how the Grievance Dictionary categories correlated with existing LIWC categories. Although high correlations with a gold standard dictionary may illustrate that the Grievance Dictionary is comparable to the LIWC in terms of psychometric qualities, we do not expect such a pattern because the Grievance Dictionary categories were designed to *supplement* LIWC categories and not replace them. Reported correlations serve to illustrate which other psycholinguistic concepts measured through the LIWC are related to each respective Grievance Dictionary category. The three highest correlating LIWC categories for each Grievance Dictionary category are depicted in Table 4 (full list of correlations available on OSF).

Overall, correlations were low, suggesting that the Grievance Dictionary does not measure precisely the same constructs as the LIWC. Most Grievance Dictionary categories were correlated to LIWC categories which one might expect to be psychologically related. For example, several Grievance Dictionary categories such as desperation, frustration, hate, jealousy, paranoia, and violence, were positively correlated to the LIWC category negative emotion. Frustration, hate, murder, threat, and violence were also positively related to the LIWC's anger category. These results may suggest that some LIWC categories serve as 'umbrella categories' for some in the Grievance Dictionary. That is, the LIWC can provide measures of more general concepts such as negative emotion, whereas the Grievance Dictionary is suited to give more granular measures of psychological constructs (e.g., frustration, paranoia) which fall into this overarching category.

Table 4. Correlations (with confidence interval) Grievance Dictionary and LIWC

| Category | Strongest correlating LIWC categories | | |
|---|---|---|---|
| deadline | cause: 0.12 [0.08;0.16] | relativity: 0.12 [0.06;0.18] | work: 0.12 [0.06;0.18] |
| desperation | cog. process: 0.10 [0.04;0.16] | discrepancy: 0.19 [0.11;0.28] | neg. emo.: 0.11 [0.04;0.18] |
| fixation | first person: 0.20 [0.10;0.3] | insight: 0.20 [0.10;0.29] | pronoun: 0.19 [0.07;0.31] |
| frustration | anger: 0.08 [0.05;0.11] | neg. emo.: 0.17 [0.07;0.27] | risk: 0.10 [0.06;0.14] |
| god | affect: 0.13 [0.08;0.17] | pos. emo.: 0.16 [0.13;0.18] | tone: 0.12 [0.11;0.13] |
| grievance | punct.: -0.05 [-0.07;-0.04] | nonfluent: -0.04 [-0.05;-0.02] | see: -0.04 [-0.06;-0.02] |
| hate | affect: 0.11 [0.05;0.16] | anger: 0.22 [0.11;0.33] | neg. emo.: 0.17 [0.06;0.27] |
| help | affect: 0.09 [0.05;0.14] | drives: 0.19 [0.08;0.29] | reward: 0.16 [0.08;0.23] |
| honour | affect: 0.19 [0.10;0.27] | pos. emo.: 0.25 [0.15;0.34] | tone: 0.18 [0.13;0.23] |
| impostor | insight: 0.08 [0.03;0.12] | pos. emo.: -0.07 [-0.11;-0.03] | risk: 0.18 [0.08;0.28] |
| jealousy | affect: 0.12 [0.05;0.19] | neg. emo.: 0.12 [0.08;0.16] | risk: 0.11 [0.06;0.16] |
| loneliness | punct.: -0.17 [-0.28;-0.06] | clout: -0.18 [-0.25;-0.11] | social: -0.15 [-0.20;-0.11] |
| murder | punct.: -0.11 [-0.13;-0.08] | anger: 0.15 [0.07;0.24] | risk: 0.10 [0.04;0.15] |
| paranoia | neg. emo.: 0.12 [0.06;0.18] | pos. emo.: -0.05 [-0.09;-0.02] | tone: -0.09 [-0.13;-0.05] |
| planning | present focus: 0.15 [0.11;0.18] | insight: 0.13 [0.06;0.21] | verb: 0.16 [0.10;0.21] |
| relation. | negation: -0.10 [-0.16;-0.04] | pos. emo.: 0.11 [0.06;0.16] | tone: 0.11 [0.06;0.17] |
| soldier | achieve: 0.11 [0.08;0.13] | power: 0.15 [0.07;0.23] | pers. pron.: -0.12 [-0.20;-0.05] |
| suicide | affect: 0.09 [0.06;0.12] | anxiety: 0.10 [0.04;0.16] | health: 0.09 [0.04;0.14] |
| surveillance | period: -0.06 [-0.08;-0.03] | pos. emo.: -0.05 [-0.08;-0.02] | you: 0.06 [0.02;0.09] |
| threat | anger: 0.20 [0.13;0.27] | cause: 0.10 [0.07;0.14] | risk: 0.12 [0.07;0.18] |
| violence | anger: 0.22 [0.12;0.32] | neg. emo.: 0.23 [0.09;0.38] | risk: 0.17 [0.08;0.26] |
| weaponry | nonfluent: -0.06 [-0.09;-0.02] | time: -0.05 [-0.08;-0.02] | you: -0.07 [-0.11;-0.03] |

*Note.* All correlations were statistically significant at the $p < 0.0023$ (0.05/22 categories) level.

**Part II: Dictionary validation**

The dictionary validation reported in this section serves to assess whether and how the Grievance Dictionary can be used to distinguish between different types of writing, for example neutral language and grievance-fuelled communications produced by terrorists or extremists. We first apply the Grievance Dictionary to different datasets to assess its external validity. Then, we test the performance of the dictionary in classification tasks.

***External validity***

We apply the dictionary to different datasets to test its validity in the context of grievance-fuelled writings (Table 5). Three tests are performed. First, following previous work on violent language use (Kaati, Shrestha, & Cohen, 2016), we make statistical comparisons between manifestos written by violent lone-actor terrorists, and large samples of 'control' texts retrieved from online forums and blogs.[3] Second, we perform a comparison between lone-actor terrorist manifestos and texts from the right-wing extremist forum Stormfront.[4] For the lone-actor terrorist samples, we draw 100-word excerpts from 22 manifestos resulting in a total sample of 4,572 texts. This 'chunking' is performed so that the average word count for the terrorist manifestos is more comparable to that of the neutral writings and Stormfront posts. For both tests, mean dictionary outcome values of the lone-actor terrorist manifestos are compared to the means of the control samples with an independent samples t-test. The control samples are down-sampled through bootstrapping to match the *n* of the lone-actor manifestos, with outcome measures reported as an average across 100 bootstrap iterations. We report the effect size for the difference by means of Cohen's $d$[5], in addition to the Bayes Factor (BF). The Bayes Factor is a measure of the degree to which the data are more likely to occur under the hypothesis that there is a difference in the dictionary categories between samples, compared to the hypothesis that there is no difference (Ortega & Navarrete, 2017; Wagenmakers et al., 2010). For example, a BF between above 10 would constitute strong evidence for the alternative hypothesis that there is a difference (Ortega & Navarrete, 2017).

---

[3] The sample was drawn from the Blog Authorship Corpus (Schler et al., 2006) and the Boards.ie forum dataset from the 2008 SIOC Semantic Data Competition: https://semantic-web.com/2008/08/27/boardsie-sioc-semantic-data-competition-starts-september-1st/

[4] All posts between 2012-2015 in the Stormfront dataset used in Kleinberg et al. (2020).

The third comparison is between abusive texts directed at politicians and neutral, stream-of-consciousness (SOC) essays (van der Vegt et al., 2020). For this comparison a dependent samples t-test is performed, because individual participants produced both types of text. Again, effect size $d$ and BF are reported for the difference between the two samples (note that this comparison is not based on bootstrapping). All results are reported in Table 6.

Overall, statistically significant differences were found for the majority of categories in all comparisons. In the majority of cases, the lone-actor texts scored higher on Grievance Dictionary categories than the control texts. In the first comparison with neutral texts from blogs and forums, the lone-actor manifestos scored higher on all categories except 'fixation' (denoted by a negative effect size $d$). The evidence for a difference between samples was very strong (BF > 10) in all cases. In the second comparison with Stormfront forum posts, the lone-actor manifestos scored proportionally higher (strong evidence with BF > 10) on the categories deadline, hate, honour, jealousy, murder, planning, soldier, threat, violence, and weaponry. In contrast, Stormfront posts scored proportionally higher (BF > 10) on desperation, fixation, impostor, loneliness, relationship, suicide, and surveillance. For the comparison between abusive writing and stream-of-consciousness texts, differences in favour of SOC texts (BF > 10) were found (denoted by negative $d$) for the categories deadline, desperation, fixation, frustration, god, grievance, hate, jealousy, paranoia, planning, and suicide. However, the abusive texts contained proportionally more references to honour, impostor, murder, surveillance, and violence (positive $d$).

Table 5. Corpora used for statistical tests

| Corpus | Number of texts | Mean word count (SD) |
|---|---|---|
| Lone-actor terrorist manifestos | 4,572 | 100 |
| Neutral texts from blogs and forums | 680,792 | 243 (503) |
| Stormfront posts | 461,950 | 95 (229) |
| Abusive writing directed at politicians | 789 | 121 (38) |
| Stream-of-consciousness (SOC) essays | 789 | 121 (35) |

---

[5] Cohen's $d$ expresses the magnitude of the difference after correcting for sample size. A $d$ of 0.20, 0.50 and 0.80 can be interpreted as a small, moderate and large effect, respectively (Cohen, 1988)

Table 6. Statistical test results (Effect size *d* with confidence interval and Bayes Factor)

| | *Manifestos vs. neutral* | | *Manifestos vs. Stormfront* | | *Abuse vs. SOC* | |
|---|---|---|---|---|---|---|
| | *d (bootstrapped)* | *BF* | *d (bootstrapped)* | *BF* | *d* | *BF* |
| deadline | 1.03 [1.02;1.04] | >10³ | 0.53 [0.51;0.54] | 314.43 | -0.48* [-0.61;-0.37] | **77.65** |
| desperation | 0.66 [0.64;0.68] | **478.28** | -0.13 [-0.14;-0.13] | **16.67** | -0.74* [-0.87;-0.64] | **167.96** |
| fixation | -0.13 [-0.13;-0.12] | **15.17** | -0.12 [-0.12;-0.11] | **11.58** | -0.74* [-0.88;-0.62] | **169.77** |
| frustration | 0.69 [0.67;0.70] | **497.29** | 0.10 [0.09;0.10] | 8.36 | -0.82* [-0.94;-0.71] | **197.47** |
| god | 0.91 [0.9;0.91] | **849.01** | 0.69 [0.68;0.69] | **513.72** | -0.10 [-0.22;0.02] | 0.45 |
| grievance | 0.65 [0.63;0.68] | **478.87** | 0.08 [0.08;0.09] | 4.90 | -0.85* [-0.99;-0.73] | **212.14** |
| hate | 1.41 [1.38;1.45] | >10³ | 0.35 [0.34;0.35] | **130.60** | -0.44* [-0.55;-0.34] | **65.26** |
| help | 0.82 [0.81;0.84] | **704.91** | 0.07 [0.07;0.08] | 2.19 | 0.04 [-0.08;0.15] | -2.48 |
| honour | 0.97 [0.96;0.98] | **973.33** | 0.52 [0.51;0.53] | **289.23** | 0.44* [0.29;0.58] | **67.02** |
| impostor | 1.35 [1.33;1.38] | >10³ | -0.29 [-0.29;-0.28] | **91.42** | 0.42* [0.33;0.54] | **61.75** |
| jealousy | 0.62 [0.61;0.63] | **408.01** | 0.16 [0.15;0.16] | **25.40** | -0.70* [-0.80;-0.61] | **153.66** |
| loneliness | 0.26 [0.25;0.27] | **74.22** | -0.19 [-0.2;-0.19] | **37.56** | -0.17* [-0.28;-0.06] | 7.55 |
| murder | 1.32 [1.31;1.33] | >10³ | 0.14 [0.13;0.14] | **18.56** | 0.21* [0.09;0.33] | **13.89** |
| paranoia | 0.51 [0.5;0.52] | **279.96** | -0.07 [-0.08;-0.07] | 3.09 | -0.90* [-1.01;-0.78] | **228.79** |
| planning | 1.12 [1.12;1.13] | >10³ | 0.72 [0.71;0.73] | **563.12** | -0.34* [-0.47;-0.23] | **39.72** |
| relation. | 0.83 [0.81;0.84] | **701.77** | -0.14 [-0.14;-0.13] | **18.66** | -0.11* [-0.22;-0.0068] | 1.53 |
| soldier | 1.79 [1.78;1.8] | >10³ | 0.74 [0.73;0.75] | **590.03** | -0.13* [-0.25;-0.01] | 3.75 |
| suicide | 1.21 [1.19;1.22] | >10³ | -0.26 [-0.27;-0.26] | **75.80** | -0.38* [-0.49;-0.28] | **49.52** |
| surveillance | 0.93 [0.92;0.94] | **895.95** | -0.15 [-0.15;-0.14] | **21.14** | 0.24* [0.14;0.35] | **19.52** |
| threat | 1.36 [1.34;1.37] | >10³ | 0.74 [0.73;0.75] | **577.13** | 0.17* [0.05;0.28] | 7.65 |
| violence | 1.5 [1.46;1.53] | >10³ | 0.57 [0.56;0.57] | **351.58** | 0.33* [0.2;0.45] | **36.5** |
| weaponry | 1.65 [1.64;1.67] | >10³ | 0.16 [0.15;0.17] | **24.78** | 0.10 [0.0018;0.22] | 0.88 |

*Notes.* A positive *d* denotes a higher score on the category for the lone-actor terrorist manifestos (test 1 and 2) and abusive texts (test 3). A BF above 10 (in bold) constitutes strong evidence for the alternative hypothesis.

## Classification
Previous work classified terrorist or extremist texts from neutral 'control samples' using the LIWC. We investigate whether the Grievance Dictionary can achieve similar results, or increase prediction performance when used to supplement the LIWC.

### Classification tasks
In four classification tasks, we examine whether the Grievance Dictionary and the LIWC can distinguish between:
1) Texts written by known terrorists vs. non-violent individuals
2) Texts written by known terrorists vs. non-violent extremists
3) Abusive vs. neutral texts (within-subject comparison of non-violent individuals)
4) An explorative cross-sample classification of extremist forum posts vs. non-extremist forum posts, trained on a dataset of text written by known terrorists vs. non-violent individuals.

All classification tasks are performed using a Naïve Bayes classifier. In Classification Task 1, we classify lone-actor terrorist manifesto excerpts (*n* = 4,572) versus neutral posts from blogs and forums (*n* = 680,792). The majority class of neutral posts is down-sampled to the same *n* as the manifesto sample by means of bootstrapping (100 times), to allow for a balanced classification task. Classification results are reported as an average across each of the 100 bootstrap tasks. In Classification Task 2, we classify lone-actor terrorist manifesto excerpts ($n$ = 4,572) versus Stormfront posts ($n$ = 461,950). Following the same procedure as in Task 1, the majority class of Stormfront posts is down-sampled 100 times. In classification Task 3, we perform classification for abusive vs. neutral, stream-of-consciousness writing with data from van der Vegt et al. (2020), using 789 texts per sample. Note that due to the smaller data size in Task 3 we do not perform bootstrapping, and instead opt for 80% of the sample as training data, and 20% as a test set to report performance metrics. In Classification Task 4, we exploratively train the model on lone-actor terrorist manifesto excerpts ($n$ = 4,572) and neutral posts from blogs ($n$ = 680,292), then test the model on Stormfront posts ($n$ = 500) vs. neutral forum posts ($n$ = 500) and report performance metrics of the latter. This task aims to replicate a potential real life setting in which models are trained on known previous terrorist cases, then applied to unseen online data which may contain extremist linguistic material relevant to security professionals.

*Feature sets*
Each classification task is performed using three different feature sets, to test the performance of the Grievance Dictionary, the LIWC and a combination of the two in classifying aforementioned datasets. The following feature sets are used:
   a) All 22 Grievance Dictionary categories.
   b) All psychological and social categories (N = 55) of the LIWC2015[6]. We exclude grammar categories from the LIWC such as pronouns and verbs because we are interested in the predictive ability of psychological concepts only, and grammatical categories do not appear in the Grievance Dictionary either.
   c) A combination of the Grievance Dictionary and psycho-social LIWC categories (N = 77).

*Results of classification tasks*
Performance metrics[7] for the classification tasks are reported in Table 7. Classification Task 1 shows high performance for distinguishing between lone-actor terrorist texts and neutral texts. In terms of accuracy, the best performing feature set was the combination of the Grievance Dictionary and the LIWC (Task 1c). Specificity and recall also show high values, but the precision of the model is low. This means that the best model is correct 43% of the time when it predicts that a text was written by a lone-actor terrorist. Classification Task 2 similarly shows that the Grievance Dictionary and the LIWC together (Task 2c) are best at distinguishing between lone-actor terrorist texts and Stormfront extremist forum posts. For this endeavour, precision is lower at 20%. Classification Task 3 shows near perfect classification when using both the Grievance and LIWC dictionary, with high performance for specificity, precision, and recall. The Grievance Dictionary alone predicts 78% of the cases accurately. In contrast, classification Task 4 shows how difficult it is to use training data from one sample (lone-actor manifestos and blog posts) when trying to classify data from another sample (Stormfront and neutral forum posts). For all three feature sets in Task 4, classification accuracy was around chance level, and further performance metrics were also sub-optimal.

---

[6] Including analytical thinking, clout, authentic language, emotional tone, affect words, social words, cognitive processes, perceptual processes, biological processes, core drives and needs, time orientation, relativity, personal concerns and informal speech.

[7] 1) Classification accuracy: true positive + true negatives / true positives + false positives + true negatives + false negatives, 2) Specificity: TN / TN + FP, 3) Precision: TP / TP + FP, 4) Recall: TP / TP + FN (see Sammut & Webb, 2011 for an overview).

Table 7. Classification results

| Task | Feature set | Accuracy | Specificity | Precision | Recall |
|------|-------------|----------|-------------|-----------|--------|
| 1. LA vs. neutral | a. Grievance | 0.97 | 0.97 | 0.17 | 0.94 |
| | b. LIWC | 0.98 | 0.98 | 0.26 | 0.99 |
| | c. Grievance + LIWC | 0.99 | 0.99 | 0.43 | 0.98 |
| 2. LA vs. Stormfront | a. Grievance | 0.93 | 0.93 | 0.12 | 0.92 |
| | b. LIWC | 0.91 | 0.91 | 0.10 | 1.00 |
| | c. Grievance + LIWC | 0.96 | 0.96 | 0.20 | 0.99 |
| 3. Abuse vs. neutral | a. Grievance | 0.78 | 0.69 | 0.74 | 0.87 |
| | b. LIWC | 0.98 | 0.97 | 0.97 | 1.00 |
| | c. Grievance + LIWC | 0.99 | 0.99 | 0.99 | 1.00 |
| 4. Cross-sample | a. Grievance | 0.55 | 0.10 | 0.53 | 1.00 |
| | b. LIWC | 0.54 | 0.95 | 0.72 | 0.13 |
| | c. Grievance + LIWC | 0.51 | 0.96 | 0.56 | 0.06 |

*Explaining high classification accuracies*

All in all, classification accuracies were high, with some close-to-perfect performances. Therefore, we examined feature importance for each task in order to discover whether the model was biased towards some features. The five most important features for each task are reported in Table 8. Feature importance rankings are based on a ROC curve analysis, where a cut-off for each feature is defined that maximizes true positives predictions, and minimizes false positives; a larger area under the ROC curve implies larger variable importance (Kuhn, 2008). Tables with ROC values for each feature per task are available on the Open Science Framework.

Features with high importance also showed stark differences in mean proportional dictionary scores between datasets. For example, the most important feature 'soldier' in Task 1a showed a mean score for lone-actor terrorist manifestos of $0.11$ (SD $= 0.07$), whereas neutral texts and Stormfront posts scored $0.01$ (SD $= 0.05$) and $0.02$ (SD $= 0.04$), respectively. This large difference between datasets will have contributed to the high prediction performance in this (and other) task(s) in that the classifier learned to over-rely on these features.

This pattern of feature differences also largely replicates the results observed in aforementioned Bayesian *t*-tests, where a decisive difference (BF $> 10^3$) was observed for 'soldier', among other variables. The second most important feature 'weaponry' (BF $> 10^3$), had a mean of $0.09$ (SD $= 0.06$) in lone-actor manifestos, in contrast to $0.02$ (SD $= 0.05$) and $0.03$ (SD $= 0.05$) in neutral texts and Stormfront posts, respectively. A full table of mean scores on features per dataset is available on the Open Science Framework.

The bias in the model towards specific features fortunately resulted in highly accurate classifications in the first three classification tasks, in that the model learned to over-rely on the features that were most apt at distinguishing between the two groups. This over-reliance on specific features may also explain the large drop in accuracy for Task 4, where the model was no longer able to rely on highly discriminative features in the training set to classify the test set (i.e., because the test set was drawn from a different context).

Table 8. Feature importance per task (top 5, full on OSF)

| Task | Feature set | Important features |
|---|---|---|
| 1. LA vs. neutral | a. Grievance | soldier, weaponry, violence, impostor, threat |
| | b. LIWC | analytic language, present focus, power, differentiation, work |
| | c. Grievance + LIWC | soldier, weaponry, violence, impostor, threat |
| 2. LA vs. Stormfront | a. Grievance | soldier, relationship, impostor, threat, hate |
| | b. LIWC | work, present focus, power, analytic language, differentiation |
| | c. Grievance + LIWC | soldier, relationship, impostor, threat, hate |
| 3. Abuse vs. neutral | a. Grievance | paranoia, grievance, frustration, fixation, desperation |
| | b. LIWC | clout, relativity, present focus, cognitive processes, analytic language |
| | c. Grievance + LIWC | clout, relativity, present focus, cognitive processes, analytic language |
| 4. Cross-sample | a. Grievance | soldier, weaponry, violence, impostor, threat |
| | b. LIWC | analytic language, present focus, power, differentiation, work |
| | c. Grievance + LIWC | soldier, weaponry, violence, impostor, threat |

## General discussion

In this paper, we introduced the Grievance Dictionary, a psycholinguistic dictionary for grievance-fuelled violence threat assessment. The aim of this work was to develop a dictionary which can specifically measure constructs relevant to threat assessment, and can be used for a wide variety of violence and extremism fuelled by a grievance. Furthermore, we aimed to address the limitations we identified pertaining to existing psycholinguistic dictionaries.

### Linguistic differences

Based on the validation results of the dictionary, we saw that the Grievance Dictionary can elucidate differences between threatening and non-threatening language. Differences in Grievance Dictionary categories were found between texts written by lone-actor terrorists, neutral writing, and extremist forum posts, as well as between abusive language and stream-of-consciousness writing. The evidence for these differences was strong.

It must be noted that a high score on Grievance Dictionary categories is not exclusive to threatening and violent texts. In our comparison between stream-of-consciousness essays and abusive writing, the former obtained significantly higher scores for categories such as desperation, fixation, and frustration. Therefore, it is important to note that high scores on single dictionary categories should not be interpreted as individual risk factors for violence, as they may also occur in non-violent texts. Instead, the measures should be interpreted jointly to gain an understanding of the content of a grievance-fuelled text, with particular attention paid to the highly 'violent' categories such as murder, violence, threats, and weaponry. Furthermore, the importance of Grievance Dictionary categories for distinguishing between different populations may also be context-dependent. For example, mentions of a (perceived) romantic relationship may positively predict violence in a threat directed at a public figure, while it may negatively predict violence (a 'linguistic protective factor') in an extremist text. Further research will be needed to establish and replicate differential meanings of Grievance Dictionary categories across contexts.

*Classification with the Grievance Dictionary*
The dictionary categories were also used to classify different types of writing, including terrorist manifestos and extremist forum posts, neutral and extremist forum posts, as well as abusive and neutral writing. The classification accuracy achieved in this study approximated or outperformed previous work in the violence research domain, for example in classifying lone-actor terrorist manifestos from Stormfront posts (0.96 here vs. 0.90 in Kaati et al., 2016). It must however be noted that precision (the percentage of true positives as a function of all positives) was sub-optimal and thus results need to be interpreted with caution.

In the classification tasks, large statistical differences between datasets led to highly accurate predictions. Therefore, it can be argued that the proposed Grievance Dictionary categories (in addition to the LIWC) are discriminatory and relevant to the grievance-fuelled violent language domain. Future work will be needed to ascertain whether the Grievance Dictionary will achieve acceptable performance on data for which it cannot rely on such strong feature differences (e.g., violent texts written by individuals who want to actualise their threat, vs. similarly violent texts written by those who do not plan to actualise). Furthermore, when train and test sets were drawn from different samples, classification accuracy was significantly reduced. Therefore, the Grievance Dictionary does not yet seem suitable for cross-contextual classifications. This is problematic seeing as a system used in a real-life security context may need to classify unseen texts that do not necessarily align with the training data in the system. These results suggest that future work will be needed before classification using the Grievance and LIWC dictionaries can be done in practice.

The results of the classification tasks also illustrate how the Grievance Dictionary and LIWC compare on such tasks. Although the LIWC alone achieved high accuracy on the classification tasks in this paper, the Grievance Dictionary sometimes outperformed or improved prediction performance when both dictionaries were used together. Even though performance with the Grievance Dictionary did not provide a major improvement over the LIWC, the Grievance Dictionary can provide more nuanced and specific psychological measures for grievance-fuelled language than the LIWC, which may be of particular interest to threat assessment practitioners. This benefit of the Grievance Dictionary also holds in cases where other classification features are used, for example bag-of-words models, parts-of-speech tags, word embeddings or bidirectional language models (see e.g. Figea et al., 2016; Neuman et al., 2015; van der Vegt et al., 2020). These methods (which do not rely on a dictionary), may sometimes perform better at classification, but are less explainable than a dictionary. In line with the ALGOCARE framework, a transparent system such as the Grievance Dictionary may be more suitable for the security domain in future.

*Usage of the Grievance Dictionary*
All things considered, the Grievance Dictionary shows promising results in distinguishing between different types of (non) grievance-fuelled language. The strong evidence for differences in dictionary measures suggest that the categories elicited from expert threat assessment practitioners hold value in understanding violent from non-violent language. However, although classification results were highly accurate on balance, precision was low and cross-sample classifications did not achieve high performance. In summary, the Grievance Dictionary can be used to make (statistical) comparisons between different text samples, or to gain a general picture of language use in a text sample. Although the Grievance Dictionary may achieve high performance in some classification tasks (i.e., where training and test sets are similar and show strong statistical differences between groups), we do not yet recommend using it for cross-domain classifications.

In order to improve (cross-domain) classification performance in future, models need to be trained on additional (larger) training samples, and a deeper understanding of domain-specific differences in dictionary categories will need to be gained. In previous work where prediction of life outcomes based on large datasets failed to show high performance, it was suggested that a good understanding of a phenomenon (e.g. shown through causal inference, such as the statistical differences observed in this study) does not necessarily always translate to accurate prediction (Garip et al., 2020). Accordingly, the Grievance Dictionary may be used to gain a deeper understanding of grievance-fuelled texts, but is not yet suitable for prediction of real-life outcomes.

Besides application in prediction, the Grievance Dictionary may be of practical use for other purposes in the field of threat assessment and violence research. For instance, it may be used to gain a broad understanding of large-scale online social media data on a user or platform level, or to compare an incoming threatening message to a (police) database of existing communications. Furthermore, the tool opens up the possibility of studying grievance-fuelled language in its full range, where Grievance Dictionary categories can be measured over time, for example to linguistically model processes of radicalisation or extremism over time (e.g. Kleinberg, van der Vegt, & Gill, 2020) or in response to specific events (Burnap et al., 2014; van der Vegt, Mozes, et al., 2019; Zannettou et al., 2019).

*Limitations and future work*

In the current paper, we have endeavoured to use the Grievance Dictionary to make meaningful comparisons between different types of violent and non-violent texts. Nevertheless, an important problem within the field of linguistic threat assessment persists. It is difficult to disentangle whether statistical differences emerged based on indicators for violence and non-violence or due to differences in topic. It is arguably not very difficult for the human eye or computer software to distinguish between a violent manifesto about attack planning and a blogpost about someone's hobby. Of particular importance is performing linguistic comparisons between violent texts written by individuals who enact violent deeds, and the same amount of violent texts written by individuals not planning to act violently. If and when data from known violent individuals is more widely available, it will be of great interest to assess whether and how differences in Grievance Dictionary categories emerge, as well as how classification tasks perform. Another way to remedy this problem is with more experimental research, where both threat actualisers and bluffers produce texts (e.g. Geurts et al., 2016) which can be assessed with the Grievance Dictionary.

Another limitation pertains to the construction of the dictionary. The seed words on which the dictionary categories are based were produced by human annotators who, to our knowledge, do not have violent ideations. Therefore, it may have been difficult for participants to produce words about attack planning and weaponry if they have little knowledge on the topic. We tried to somewhat ameliorate this problem by including word candidates obtained through automatic methods. Nevertheless, future improvements to the Grievance Dictionary may include word candidates that are obtained by means of a data-driven approach. That is, we may extract words from texts which are known to have been written by lone-actor terrorists or other violent individuals to serve as seed words.

Lastly, the assumption that the Grievance Dictionary categories indeed measure the (psychological) constructs they are designed to measure remains to be tested. For example, we do not yet know whether someone who is experiencing jealousy will also use more words from the jealousy category in the dictionary. This limitation holds for many psycholinguistic dictionaries including the LIWC, and highlights the importance of obtaining ground truth emotion datasets (Kleinberg, van der Vegt, & Mozes, 2020). Alternatively, emotions (and potentially other psychological constructs) can be experimentally manipulated prior to text writing in order to ascertain that the true emotional state of the text author is inferred from text (Kleinberg, 2020; Marcusson-Clavertz et al., 2019). Therefore, future work on the Grievance

Dictionary and other psycholinguistic dictionaries should focus on measuring or even eliciting psychological processes such as frustration, jealousy, and loneliness, then measuring whether these constructs also emerge in language when applying the Grievance Dictionary.

**Conclusion**

The purpose of the Grievance Dictionary is to serve as a resource for threat assessment practitioners and researchers aiming to gain a better understanding of grievance-fuelled language use. Initial validation tests of the dictionary show that differences between violent and non-violent texts indeed can be detected and classified using the dictionary. All information regarding the construction and specifications of the dictionary is available to researchers and practitioners, so that the capabilities *and* limitations of the Grievance Dictionary can be adequately scrutinised. Even though future research will be needed to ascertain the utility of the dictionary in other contexts (such as violent texts from authors with no violent intent), we hope the current work serves as an impetus to gain a better understanding of grievance-fuelled language by automatic means.